\documentclass[times, 10pt,twocolumn]{article}
\usepackage{latex8}
\usepackage{times}
\usepackage{amsmath}
\usepackage{algorithm}
\usepackage{algorithmic}
\usepackage[T1]{fontenc}
\usepackage[pdftex]{graphicx, color}

\begin{document}

\title{An Empirical Study of the Effects of Sample-Mixing Methods for Efficient Training of Generative Adversarial Networks}

\author{Makoto Takamoto, Yusuke Morishita,  \\
Biometrics Research Laboratories, NEC \\ 
1753 Shimonumabe, Nakahara-ku, Kawasaki-shi, Kanagawa, Japan\\ makoto.takamoto@nec.com, \\
}

\maketitle
\thispagestyle{empty}

\begin{abstract}
  It is well-known that 
  training of generative adversarial networks (GANs) requires huge iterations 
  before the generator's providing good-quality samples. 
  Although there are several studies to tackle this problem, 
  there is still no universal solution. 
  In this paper,
  we investigated the effect of sample mixing methods, that is, Mixup, CutMix, and newly proposed Smoothed Regional Mix (SRMix), 
  to alleviate this problem. 
  The sample-mixing methods are known to enhance the accuracy and robustness in the wide range of classification problems, 
  and can naturally be applicable to GANs
  because the role of the discriminator can be interpreted as the classification between real and fake samples.
  We also proposed a new formalism applying the sample-mixing methods to GANs 
  with the saturated losses 
  which do not have a clear "label" of real and fake. 
  We performed a vast amount of numerical experiments using LSUN and CelebA datasets. 
  The results showed that Mixup and SRMix improved the quality of the generated 
  images in terms of FID in most cases,
  in particular, SRMix showed the best improvement in most cases.
  Our analysis indicates that 
  the mixed-samples can provide different properties 
  from the vanilla fake samples, 
  and the mixing pattern strongly affects the decision of the discriminators.
  The generated images of Mixup have good high-level feature but 
  low-level feature is not so impressible. 
  On the other hand,
  CutMix showed the opposite tendency. 
  Our SRMix showed the middle tendency, that is, showed good high and low level features.
  We believe that 
  our finding provides a new perspective to accelerate the GANs convergence
  and improve the quality of generated samples. 
\end{abstract}

\section{INTRODUCTION}

Generative adversarial networks (GANs) \cite{goodfellow2014generative} are
considered as one of the most promising frameworks for data-synthesis. 
Although there are some variations, 
the training of GANs is an adversarial game between two neural networks: 
one is a generator
which tries to synthesize realistic sample (fake samples), 
and the other is a discriminator
which tries to distinguish between real and fake samples. 

Despite its promising properties of GANs, 
it is also known that GANs training is difficult \cite{odena2019open}. 
To make matters worse,
it demands huge iterations before providing good-quality samples.
Concerning the difficulty of GANs training,
a considerable amount of effort has been conducted on finding its origin
and several regularization techniques have been found effective for stabilization
\cite{roth2017stabilizing,kodali2018on}.
In particular, 
one of the most accepted findings is that 
training a good discriminator is a key for a stable and fast training of GANs 
\cite{NIPS2017_7159,miyato2018spectral,zhang2020consistency}.
In spite of those research activities, 
this is still an open problem for the GANs research community. 

In this paper,
we investigated the effect of sample mixing methods, that is, Mixup \cite{zhang2018mixup}, CutMix \cite{yun2019cutmix},
and newly proposed Smoothed Regional Mix (SRMix) to alleviate the above problem.
The sample-mixing methods enhance the accuracy and robustness in the wide range of classification problems. 
This means that the sample-mixing methods can be expected to effective also for GANs 
because the role of the discriminator can be interpreted as the classification between real and fake samples. 
We proposed a new formalism applying the sample-mixing methods to GANs training, 
in particular, GANs with modern saturated losses such as 
the Wasserstein loss \cite{pmlr-v70-arjovsky17a} and Hinge loss \cite{lim2017geometric}
which do not have a clear "label" of real and fake samples.
In addition, 
our simple implementation allows us to combine it with any regularization methods, 
such as gradient penalty \cite{NIPS2017_7159}, spectral-normalization \cite{miyato2018spectral},
and consistency-regularization \cite{zhang2020consistency}. 
We performed a vast amount of numerical experiments using LSUN and CelebA datasets, 
and analyzed the effects of sample-mixing methods on GANs training. 

In summary, 
our contributions are as follows:
\begin{itemize}
    \item
    We proposed a new sample-mixing method: Smoothed Regional Mix (SRMix).
    We also proposed a new formalism of applying the sample-mixing methods to GANs
    with saturated loss functions.
    (Section \ref{sec:method}). 
  \item We performed comprehensive numerical experiments,  
    and found that Mixup and SRMix were effective also for GANs. 
    (Section \ref{sec:experiments}). 
  \item We analyzed the resulting fake samples,
    and provided the insights of the effects of each sample-mixing method 
    (Section \ref{sec:analysis}). 
\end{itemize}

\section{Related Work}

\textbf{GANs} \quad Generative adversarial networks (GANs) \cite{goodfellow2014generative} are
considered as one of the most promising frameworks for data-synthesis.  
One of the most attractive properties of GANs is their flexibility; 
they allow us to synthesize nearly any kind type of the data, 
such as image generation \cite{karras2018progressive,zhang2019self,brock2019large,karras2019a}, 
super-resolution of images \cite{ledig2017photo}, 
text \cite{hu2017toward,yu2017seqgan}, 
voice \cite{hono2019singing}, 
and even text-to-image \cite{reed2016generative}. 

Concerning the difficulty of GANs training,
although this is still a very hot research topic, 
there are several techniques that are known very effective for stabilization, 
such as the gradient penalty \cite{NIPS2017_7159},
and spectral-normalization \cite{miyato2018spectral}.
Recently
it was found that the consistency-regularization \cite{sajjadi2016regularization,laine2017temporal},
which is one of the most popular methods for semi-supervised learning, 
is very effective for the stabilization of GANs training \cite{zhang2020consistency}. 
Our paper followed this line, 
and analyzed another popular semi-supervised methods: Mixup and its variants. 

\textbf{Mixup} \quad Mixup \cite{zhang2018mixup} is a regularization method for DNN classifiers 
by creating virtual samples and labels in the vicinity of the distributions of the mixed data
\cite{guo2019mixup}. 
This method is known effective for many problems of DNN, 
such as memorization, sensitivity to adversarial examples,
and supporting semi-supervised learning \cite{berthelot2019mixmatch}. 
\cite{verma2019manifold} found that interpolating hidden states can result in a better representation. 
In spite of this success,
Mixup is also known to introduce unnatural artifacts because of the global mixing, 
leading sub-optimal performance of the classifiers. 
CutMix \cite{yun2019cutmix} is a method 
which alleviate this problem by creating a new sample by regionally mixing two samples. 
The authors found that 
this regional mixing encourages DNN classifiers to focus on discriminative local parts, 
resulting in consistent performance gains. 
Note that in \cite{zhang2018mixup}, the authors tried to apply Mixup 
to the original GAN loss (non-saturated loss), and reported the training were stabilized. 
However, the authors reported neither Inception nor FID score.
In addition,
we also emphasize that
the author's method cannot directly be applied to the saturated losses 
because they do not have a clear "label" of real and fake samples 
which is one of the key components of Mixup.
In this paper,
we applied the sample-mixing methods for various saturated loss functions,
and analyzed their effects. 

\section{Methods}
\label{sec:method}
\begin{algorithm}[t]
  \caption{Algorithm of GANs training with mixed-samples}
  \label{alg1}
  \textbf{Input} generator and discriminator parameters $\theta_{\rm G}, \theta_{\rm D}$, ladder ratio r, Adam hyper-parameters $\eta$,
    $\beta_1, \beta_2$, batch size $M$, number of discriminator iterations per generator iteration $n_{\rm crit}$
  \begin{algorithmic}[1]
    \FOR{number of training iterations}
    \FOR{t $= 1$ \TO $n_{\rm crit}$}
    \FOR{i $= 1$ \TO $M$}
    \STATE Sample real data $x \sim p_{\rm data}(x)$, latent variable $z \sim p(z)$, 
    \IF{$i \leq r M$}
    \STATE $\tilde{x} \leftarrow$ mixed-sample
    \ELSE
    \STATE $\tilde{x} \leftarrow G(z)$
    \ENDIF
    \STATE $L_{\rm D}^{(i)} \leftarrow D(\tilde{x}) - D(x)$
    \ENDFOR    
    \STATE $\theta_{\rm D} \leftarrow$ Adam $(\sum_{i=1}^M L_{\rm D}^{(i)}, \eta, \beta_1, \beta_2)$      
    \ENDFOR
    \STATE Sample a batch of latent variables $\{ z^{(i)} \}^M_1 \sim p(z)$
    \STATE $\theta_{\rm G} \leftarrow$ Adam $(\sum_{i=1}^M (- D(G(z))), \eta, \beta_1, \beta_2)$      
    \ENDFOR
  \end{algorithmic}
\end{algorithm}

In this section,
we provide a detailed explanation of our new formalism to combine sample-mixing methods 
to GANs with saturated loss functions effectively.
As is well-known, 
Mixup assumes to mix both samples and labels
which is necessary to encourage classifiers to learn the mixed-samples as
located in the vicinity of the distributions of the mixed data. 
However,
GANs with the saturated loss functions do not have "label" of real and fake samples,
so that Mixup cannot directly be applied in this case.
It is also non-trivial how to provide the mixed samples to the discriminator
without interfering the discriminator's learning of the fake samples. 
To solve these problems,
we proposed the following new formalism. 

In our formalism,
we proposed not to use "label" mixing 
because of the absent of the label in the case of the saturated loss functions
\footnote{%
  Note that it is possible to use the generator's output of real and fake samples
  as a pseudo-label.
  However, our numerical experiments using the pseudo-label did not show any
  improvement of FID score. 
}%
.
Instead,
our formalism regards the mixed-sample as a new kind of fake samples
whose distribution is located in the vicinity of the distributions of real and fake samples
but not necessarily between them. 
In the case of the standard procedure for stable GANs training, 
it is common to construct different mini-batches for real and fake samples.
In our formalism,
a certain amount of fake samples were replaced by the mixed-samples 
at every iteration of discriminator's training. 
This allows us to preserve a certain amount of the original fake data distribution 
in terms of the discriminator evaluation,
which is necessary for effective training of the generator
by the discriminator's evaluation. 
A detailed explanation of the flow is given in Algorithm \ref{alg1}. 

\begin{table}[t]
  \centering
  \caption{A summary of training setups in Section \ref{sec:ex3}.
    NM$_D$ means the normalization of the discriminators.
    LN means the layer normalization,
    and SN the spectral normalization.
    GP means the gradient penalty.
    CR means the consistency regularization. 
    }
  \begin{tabular}{c|c|c|c|c} \hline
    Names  & Model  & Loss  & NM$_D$   & Regularization  \\ \hline    
    Case 1 & DCGAN  & Hinge & LN       & CR \& GP \\ 
    Case 2 & DCGAN  & Hinge & SN       & CR       \\ 
    Case 3 & ResNet & Hinge & LN \& SN & CR \& GP \\ \hline
    Case 4 & StyleGAN2 & \multicolumn{3}{|c}{vanilla implementation} \\ \hline        
  \end{tabular}
  \label{table:table_models}
\end{table}
\begin{table*}[t]
  \centering
  \caption{FID score of the GAN training with the help of sample mixing. 
    }
  \begin{tabular}{c|c|c|c|c|c|c} \hline
    Case & Type     & Bedroom & Church & Bridge & Tower & CelebA \\ \hline
    Case 1 &vanilla & 22.2 $\pm$ 4.2 & 15.5 $\pm$ 2.3 & 22.1 $\pm$ 1.5 & {\bf 15.1 $\pm$ 2.2} & {\bf 11.5 $\pm$ 0.7} \\
           &Mixup   & 20.3 $\pm$ 3.4 & 14.8 $\pm$ 2.2 & 23.4 $\pm$ 1.8 & 16.5 $\pm$ 2.1 & {\bf 11.6 $\pm$ 0.9}  \\ 
           &CutMix  & 22.2 $\pm$ 1.2 & {\bf 13.6 $\pm$ 0.6} & 25.6 $\pm$ 5.2 & 16.6 $\pm$ 1.5 & 12.4 $\pm$ 0.7  \\ 
           &SRMix   & {\bf 19.5 $\pm$ 1.6} & {\bf 13.4 $\pm$ 1.2} & {\bf 21.2 $\pm$ 2.9} & {\bf 15.0 $\pm$ 1.3} & 12.2 $\pm$ 0.6 \\ \hline
    Case 2&vanilla & 33.9 $\pm$ 1.8 & 20.4 $\pm$ 2.1 & {\bf 27.0 $\pm$ 1.7} & 21.8 $\pm$ 2.5 & 15.5 $\pm$ 0.7  \\
          &Mixup   & 30.7 $\pm$ 3.7 & {\bf 17.8 $\pm$ 0.5} & 28.6 $\pm$ 1.3 & 20.8 $\pm$ 1.4 & 14.1 $\pm$ 0.6  \\ 
          &CutMix  & 37.8 $\pm$ 1.2 & 24.2 $\pm$ 1.3 & 36.8 $\pm$ 3.2 & 26.6 $\pm$ 1.5 & 18.0 $\pm$ 0.8  \\ 
          &SRMix   & {\bf 27.0 $\pm$ 2.5} & 18.4 $\pm$ 1.3 & {\bf 27.1 $\pm$ 1.7} & {\bf 19.7 $\pm$ 1.2} & {\bf 13.8 $\pm$ 1.0}  \\ \hline    
  \end{tabular}
  \label{table:tableM}
\end{table*}
In this paper, 
we considered the following sample-mixing methods: 
Mixup \cite{zhang2018mixup}, 
CutMix \cite{yun2019cutmix},
and newly proposed Smoothed Regional Mix (SRMix). 
The creation of a mixed samples by those method can formally be written as:
\begin{align}
  \tilde{x} = \mathbf{M} \odot x_i + (\mathbf{1} - \mathbf{M}) \odot x_j, 
\end{align}
where $x_i$ and $x_j$ are two sample data,
$\mathbf{M} \in [0,1]^{W \times H}$ is a mask,
and $\odot$ is element-wise multiplication. 
In the case of Mixup,
the mask $\mathbf{M}$ is constant for all the images,
and its value (described as $\lambda$ in the paper \cite{zhang2018mixup}) 
is sampled from the Beta distribution as $\lambda \sim \mathrm{Beta}(\alpha, \alpha)$,
for $\alpha \in (0,\infty)$. 
In the case of CutMix,
the mask $\mathbf{M}$ has a rectangular non-zero (unity) region
whose box coordinates $(r_x,r_y, r_w, r_h)$ are determined by uniform sampling.
In the case of SRMix, 
we regionally mix two samples as CutMix but allows a transient region 
by connecting them by a smooth function such as the hyperbolic-tangent function:
the mask $\mathbf{M}$ can be written as:
\begin{align}
  \mathbf{M} \equiv \frac{1}{2} [1 + \sigma \tanh((x^i - x^i_0)/\Delta x^i)], 
\end{align}
where $\sigma$ is either $\pm 1$ determined randomly, 
and $x^i$ denotes one of the pixel coordinate
(either horizontal or vertical direction) 
whose direction is randomly determined for each sample.
$x^i_0$ is the central coordinate of the transient region,
and $\Delta x^i$ is the width of the transient region
\footnote{%
  Recently a similar method, SmoothMix \cite{lee2020smoothmix}, was proposed
  which is CutMix with soft edge for classification problems. 
  We found that our SRMix corresponds to a simpler version of this methods. 
}%
.
Note that 
SRMix not only simulates CutMix but Mixup by introducing the transient region. 
In the following experiments,
$\alpha$ of the Beta distribution is set to unity, that is,
uniform distribution of sampling is assumed; 
$x^i_0$ is chosen from 1/8 to 7/8 of the image coordinate,
and $\Delta x^i$ ranges from 2 pixels to 1/16 of the resolution of the image.


\section{Experiments}
\label{sec:experiments}

In this section,
we provided the numerical results of our experiments
to investigate the effects of the sample-mixing methods on GANs training. 
We performed the experiments using several GAN architectures and loss functions
on five datasets. 
All the scores were measured by Frechet Inception Distance (FID) \cite{heusel2017gans}. 
In particular,
we used all the training data as the reference of real samples when calculating FIDs, 
and compared them with 10k fake samples. 
In all the tests,
we performed five-training and obtained averages and standard deviations 
to reduce statistical fluctuations.
The training was performed using PyTorch 1.3 for Case 1-3 and Pytorch 1.6 for Case 4 \cite{NEURIPS2019_9015}. 

\textbf{Datasets} \quad We evaluated our method using five datasets,
CELEBA-HQ-128 (CelebA) \cite{liu2015faceattributes},
LSUN's training data of bedroom, church\_outdoor, tower, and bridge \cite{yu15lsun}. 
On all the dataset,
we performed Resize, CenterCrop, and Normalize using PyTorch APIs in this order. 
The resolutions are $128 \times 128$ on CelebA in Section \ref{sec:ex3}, 
and $64 \times 64$ on the other cases. 

\textbf{Models and Loss functions}
\quad
As model architectures of the generator and discriminator, 
we used DCGAN \cite{radford2016unsupervised} and 
the ResNet-like structure proposed in \cite{NIPS2017_7159}.
On the generator,
batch normalization \cite{pmlr-v37-ioffe15} was used. 
On the discriminator,
either spectral normalization or layer normalization \cite{ba2016layer}, 
or both, was used. 
In the Case 4, we used StyleGAN2 \cite{Karras2019stylegan2}
\footnote{
    To perform the experiment for StyleGAN2,
    we used an implementation provided in 
    https://github.com/lucidrains/stylegan2-pytorch
    whose tag is 1.5.6, with its vanilla setting and parameters.
    We really appreciate the authors of the repository. 
}.
The summary of the models, loss functions, normalization of the discriminator,
and regularization are listed in Table I 
\footnote{
  In the cases 1 and 3 of Section \ref{sec:ex3},
  gradient penalty was used for every 5 iterations,
  which we found was enough for improving FID scores
  and allowed us a more efficient training. 
}. 
For simplicity,
we used unconditional GANs for all the cases. 

\textbf{Training}
\quad
For the training, 
we set the batch size 64 in the Cases 1-3,
and 4 in the Case 4. 
  The training was stopped after 100k iterations in Cases 1-3 and 150k iterations in Case 4.
Note that the number of iterations was relatively small. 
This is because the deformation of image contents created by sample mixing
can be expected harmful to discriminators training in the final-phase, 
which will be discussed in Sections \ref{sec:ex3} and \ref{sec:analysis}. 
The optimization was performed using Adam \cite{kingma2014adam}
with parameters $(\beta_1, \beta_2) = (0.01, 0.999)$.
The learning rate was set to $\eta = 10^{-3}$ in all the generators 
and the discriminator in Section \ref{sec:ex3}. 
The number of discriminator iterations per generator iteration was set as
$n_{\rm crit} = 2$.
Concerning the consistency-regularization,
the augmentation is a combination of randomly flipping the image horizontally
and randomly shifting the image by 4 pixels 
following \cite{zhang2020consistency}.
The coefficient of consistency regularization was set as unity in all the experiments
in Section \ref{sec:ex3}. 

\section{\label{sec:ex3}Results}
\begin{table*}[t]
  \centering
  \caption{Development of FID scores of the GAN training in terms of the iteration numbers 
    in the Cases 3 using LSUN bedroom dataset. 
    }
  \begin{tabular}{c|c|c|c|c|c|c} \hline
    Type    & 1k & 10k & 25k & 50k & 75k & 100k \\ \hline
    vanilla &210.6 $\pm$ 26& 70.7 $\pm$ 45.7 & {\bf 22.5 $\pm$ 4.6} & 13.1 $\pm$ 2.7 & 12.0 $\pm$ 1.7 & 9.4 $\pm$ 1.2  \\
    Mixup   &210.0 $\pm$ 24& 80.0 $\pm$ 53.8 & 25.5 $\pm$ 10.5 & 13.5 $\pm$ 2.7 & 10.3 $\pm$ 2.3 & {\bf 8.7 $\pm$ 1.2}  \\ 
    CutMix  &214.3 $\pm$ 21& {\bf 59.4 $\pm$ 15.6} & 24.5 $\pm$ 7.2 & 13.2 $\pm$ 3.2 & 11.2 $\pm$ 2.6 & 9.4 $\pm$ 0.6  \\ 
    SRMix   &{\bf 206.1 $\pm$ 18}&  76.0 $\pm$ 42.5 & {\bf 22.8 $\pm$ 4.2} & {\bf 11.6 $\pm$ 1.7} & {\bf 9.4 $\pm$ 1.0} & 9.7 $\pm$ 1.0  \\ \hline    
  \end{tabular}
  \label{table:tableM2}
\end{table*}
\begin{figure*}
  \centering
  \includegraphics[bb= 0 100 2304 1296,width=15cm]{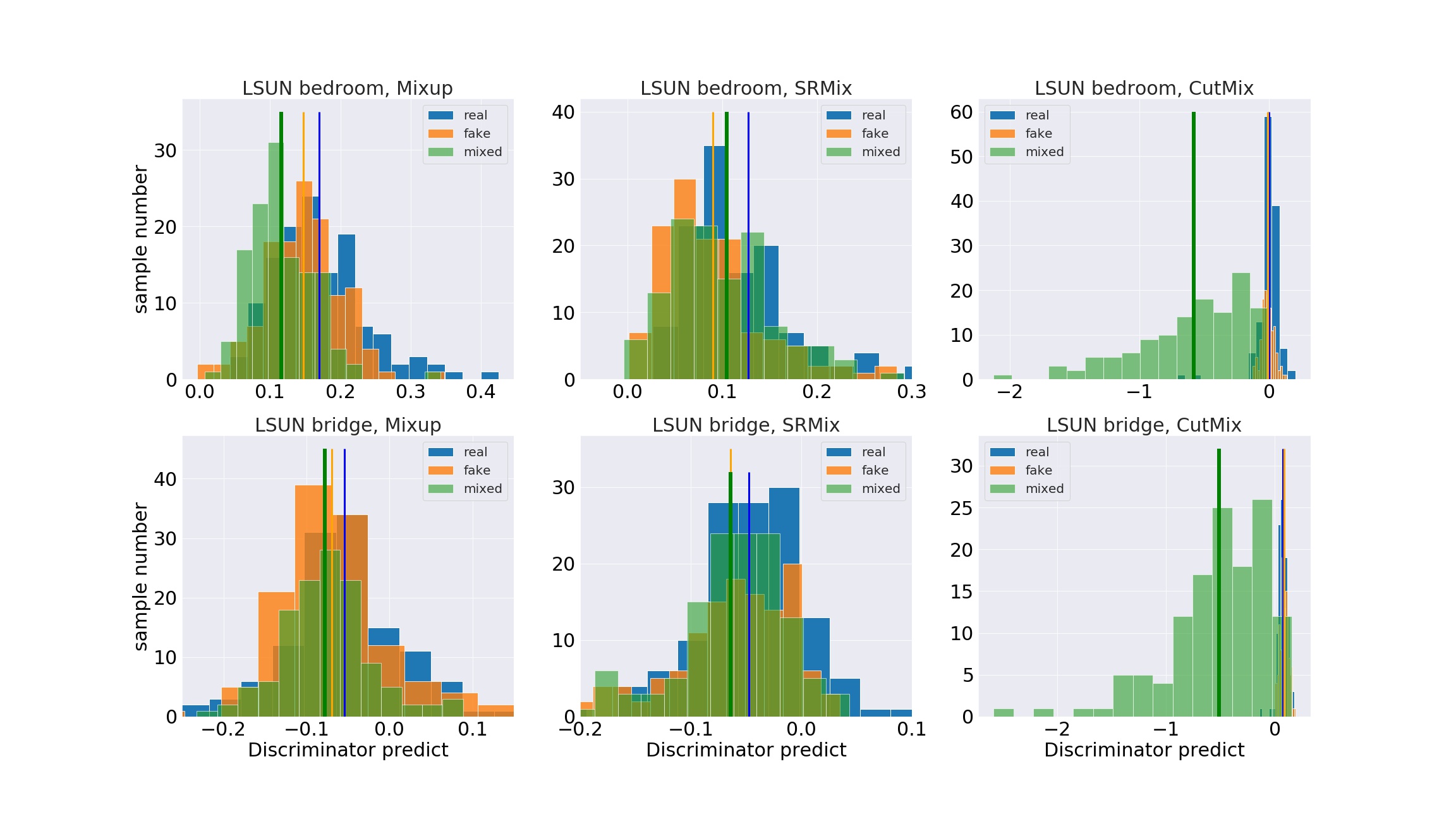}    
  \label{fig:figDMixLSUNbridge}
  \caption{Distribution functions of real, fake, and mixed samples
    created by Mixup, CutMix, and SRMix in Case 2. 
    Top panels: LSUN bedroom dataset. 
    Bottom panels: LSUN bridge dataset.    
    The blue, orange, and green histograms corresponds with real, fake, 
    and mixed samples' discriminator evaluation, respectively.
    }
\end{figure*}
In this section,
we present the results of the experiments using Mixup, CutMix, and SRMix. 
The mixed samples occupied 25 \% in the Cases 1 and 2, and 15 \% in the Cases 3 and 4 
in one mini-batch, respectively. 
The results were listed in Table II 
which showed that adding mixed samples mostly improved FID scores comparing with vanilla training. 
In particular,
they indicated that SRMix worked well in most cases;
On the other hand,
CutMix failed to improve FID in most cases. 
To understand this tendency,
in top-panels in Figure 1, 
we plotted the distributions of the discriminator's evaluation of the quality of
real, fake, and mixed samples
in the Case 1 using LSUN bedroom dataset 
where Mixup and SRMix showed improvement of their FID scores. 
It showed that 
the location of the mixed sample distribution correlated with FID scores. 
In the cases of Mixup and SRMix,
the created mixed samples are distributed around fake samples,
in particular, samples created by SRMix are distributed between real and fake samples. 
On the contrary,
the mixed samples by CutMix are distributed much worse than fake samples. 
On the other hand,
bottom-panels in Figure 1 
is the plot of the sample distributions
of real, fake, and mixed samples in the case of Case 2 using LSUN bridge
where all the methods failed to improve their FID scores. 
It showed that 
the mixed sample distributions of all the cases failed to produce better samples 
than the fake ones, indicating the importance of the better mixed samples than fake ones.
The above results indicate that 
the FID scores became better
when the mixed samples are distributed between real and fake samples. 
Note that this is the intended behavior of sample-mixing methods 
but interestingly this did not always work well. 
In Section \ref{sec:analysis} we analyze the reason of this behavior more deeply. 

Table III showed the development of FID scores in the Case 3
using LSUN bedroom dataset. 
It showed that 
the sample-mixing did not always work in the early phase.
Besides, 
SRMix worked well around the middle-phase (50k to 75k iterations) but showed a poor performance 
in the late phase (100k). 
We consider that this is because 
the samples created by SRMix was fruitful for the discriminator around the middle-phase
but became too easy at the late-phase 
because of the ability of ResNets's capturing the high-level feature of images
which can detect unnatural artifacts introduced by the sample-mixing. 

\begin{table}[t]
  \centering
  \caption{The average and minimum value of the FID scores of the GAN training
           in the Cases 4 using CelebA dataset. 
    }
  \begin{tabular}{c|c|c} \hline
    Case    & average & minimum \\ \hline
    vanilla & {\bf 20.7 $\pm$ 1.2} & 19.4 \\
    Mixup   & 22.0 $\pm$ 2.4 & 19.9 \\
    CutMix  & 23.6 $\pm$ 1.4 & 21.6 \\
    SRMix   & 22.6 $\pm$ 2.5 & {\bf 19.1} \\ \hline    
  \end{tabular}
  \label{table:table_SG2}
\end{table}

\begin{figure*}
  \centering
  \includegraphics[width=15cm]{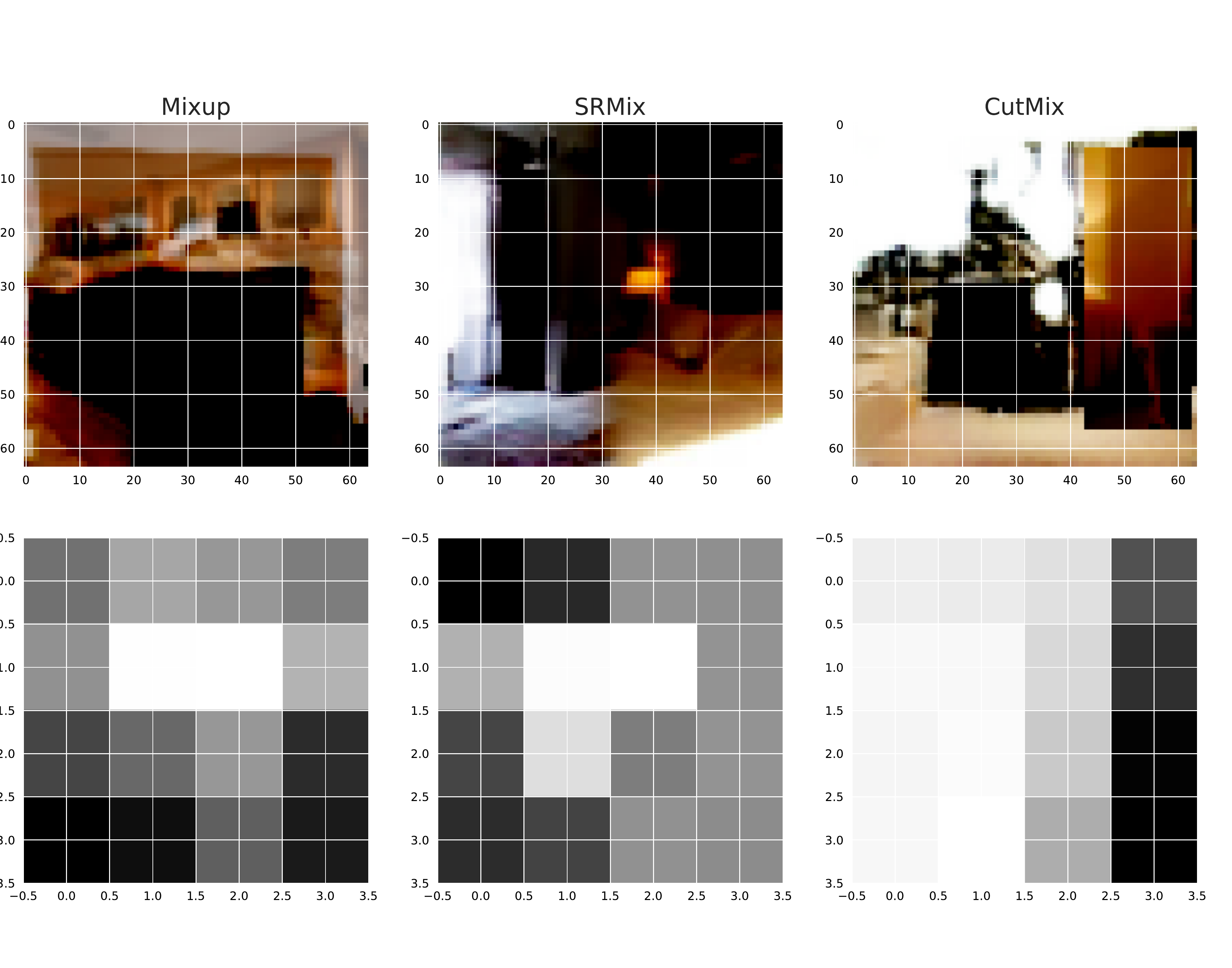}    
  \label{fig:figMixes}
  \caption{
    Top panels are the examples of Mixed-samples generated by
    Mixup (left), SRMix (middle), and CutMix (right). 
    The bottom panels are the corresponding discriminator's outputs
    in $4 \times 4$ regions (the darker, the worse). 
    SRMix image (middle-top) has a transient region around the middle of the image.
    CutMix image (right-top) has a region in the right-hand side of the image
    where fake sample is cutting-in. 
    }
\end{figure*}
In the Case 4,
we performed experiments using StyleGAN2
which is known as one of the most successful GANs model at present,
allowing us to generate high-resolution images. 
Table IV showed the average and minimum FID scores, 
using CelebA bedroom dataset.
Similar to the Case 3, 
it also indicated that 
the sample-mixing did not work well in this case. 
We consider that this is due to the ability of StyleGAN2 capturing
both the low and high-level feature of images. 
However,
we also noted that
SRMix resulted in the best FID value in terms of the minimum value
of the obtained FID in 5 trials. 
This indicates that 
sample-mixing methods can be beneficial even for modern GAN models
capable of capturing high-level features,
but needs a more sophisticated methods to control the strength of the
fluctuation from sample-mixings, such as adaptively changing sample-mixing
samples ratio in fake data. 

\section{Analysis}
\label{sec:analysis}
\begin{figure*}
  \centering
  \includegraphics[width=12cm]{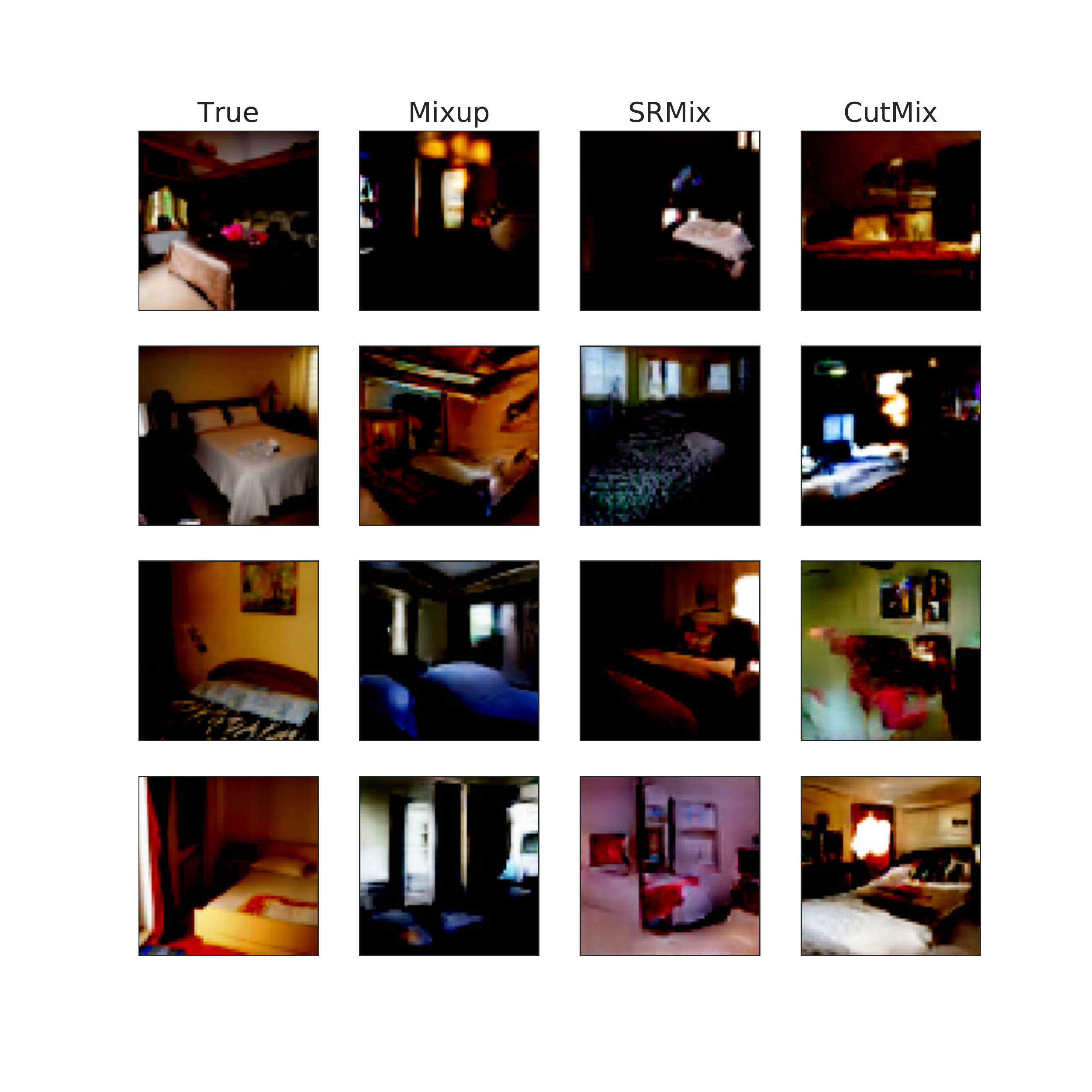}
  \label{fig:all_img}
  \caption{Comparison of true and fake images generated by the generators
           trained using Mixup, SRMix, and CutMix. 
    }
\end{figure*}

In this section,
we analyzed the effects of mixed-samples on the discriminator's decisions. 
Figure 2 is the generated images by sample-mixing (Top)
and the discriminators' outputs\footnote{
  Note that the discriminator's outputs were normalized to range from 0 (black)
  to unity (white) by the post-process. 
} (Bottom).
We found that the discriminator of CutMix properly detected the region
where the fake sample is cut into the real sample (right-hand side of the image). 
On the contrary,
the discriminator of SRMix did not react to the transient region (middle of the image). 
This indicates that
the discriminator of CutMix used the strong edge of the boundary region
as a clue to find sample-mixing samples. 
This was also be indicated from the left-hand side of the discriminator's output
where the discriminator's judge becomes white, meaning real sample, in almost
all the region, showing that the discriminator neglects other image features
to judge if the sample is real or fake.
Concerning the Mixup (left panels), 
it showed that
the discriminator paid attention to various regions
because of the global linear interpolation.
However,
it failed to give a penalty on the unnatural regions 
resulting from the linear interpolation of two images,
which does not occur in the real images. 
On the other hand,
SRMix allowed the discriminator to learn from the true real and fake samples 
avoiding appearance of the strong edge,
which makes SRMix as a better provider of samples between real and fake
in most cases. 

Figure 3 plots the true and fake images generated by the generators
trained using Mixup, SRMix, and CutMix.
This indicates that
Mixup encouraged the generator to produce images with good high-level information,
but the low-level information (e.g. form of bed) is relatively poor.
On the other hand, 
CutMix encouraged the generator to produce images with good low-level information,
but the high-level information is poor.
SRMix showed the intermediate features of the two methods.
We consider that
those tendency played an important role for a better FID score of Mixup and SRMix
because FID measures the distance of high-level information
between real and fake images. 

\section{\label{sec:discussion}Discussion And Conclusion}

In section \ref{sec:experiments},
we performed the numerical experiments using mixed samples to GANs training. 
In most cases, we observed the improvement of the FID score,
in particular, in the case of SRMix. 
However, 
the improvement of the FID scores was relatively unstable 
and sometimes became even worse than vanilla training.
One of the reasons for this was indicated in Figure 1 
which showed that
the produced samples were not always located between real and fake samples 
but located far left of fake samples,
indicating too easy for discriminators to judge as fake samples. 
We consider that this can be partly due to the deformation of the contents in
the resulting samples, for example, human faces and buildings,
which can be too easy for a well-trained discriminator to detect. 
This means that
we may have to stop using mixed-samples in the later phase of the GANs training 
since this is an intrinsic problem of the sample-mixing method. 

In conclusion, 
the sample-mixing methods were indicated to be fruitful even for 
an effective GANs training from our numerical experiments. 
On the other hand,
it was also shown that
our proposed methods to create mixed-samples did not always work,
in particular, when the samples failed to be located between
real and fake data in terms of the discriminator's evaluation. 
In order for a better GAN training, 
it is crucial to find a method to create good samples 
more stably than Mixup, CutMix, and SRMix, 
and we will tackle this in our future work. 

\section*{Acknowledgment}
We would like to thank our team members 
for many fruitful comments and discussions.
We also would like to thank our anonymous referees 
for their fruitful comments. 

\bibliographystyle{latex8}

\end{document}